\documentclass[10pt,twocolumn,letterpaper]{article}

\usepackage{cvpr}
\usepackage{tikz}
\usepackage{times}
\usepackage{epsfig}
\usepackage{graphicx}
\usepackage{amsmath}
\usepackage{amssymb}

\usepackage[hyphens]{url}

\usepackage[pagebackref=true,breaklinks=true,letterpaper=true,colorlinks,bookmarks=false]{hyperref}

\newcommand{\Xa}[1]{\bar{X}^{#1}_{\text{train}}}

\newcommand{\Xe}[1]{\bar{X}^{#1}_{\text{test}}}
\newcommand{\NN}[2]{{#2}^{(#1)}}
\newcommand{\RELU}[2]{\text{ReLU}^#1_#2}
\cvprfinalcopy % *** Uncomment this line for the final submission

\def\cvprPaperID{2121} % *** Enter the CVPR Paper ID here

\usepackage{breakurl}
\ifcvprfinal\fi
\begin{document}

%%%%%%%%% TITLE
\title{Building a Regular Decision Boundary with Deep Networks}

\author{Edouard Oyallon\\
D\'epartement Informatique\\
Ecole Normale Sup\'erieure\\
Paris, France\\
{\tt\small edouard.oyallon@ens.fr}
}%\and

\maketitle
%\thispagestyle{empty}

%%%%%%%%% ABSTRACT
\begin{abstract}
In this work, we build a generic architecture of Convolutional Neural Networks to discover empirical properties of neural networks. Our first contribution is to introduce a state-of-the-art framework that depends upon few hyper parameters and to study the network when we vary them. It has no max pooling, no biases, only 13 layers,  is purely convolutional and yields up to 95.4\% and 79.6\% accuracy respectively on CIFAR10 and CIFAR100. We show that the nonlinearity of a deep network does not need to be continuous, non expansive or point-wise, to achieve good performance. We show that increasing the width of our network permits being competitive with very deep networks. Our second contribution is an analysis of the contraction and separation properties of this network. Indeed, a 1-nearest neighbor classifier applied on deep features progressively improves with depth, which indicates that the representation is progressively more regular. Besides, we defined and analyzed local support vectors that separate classes locally.  All our experiments are reproducible and code is available online, based on TensorFlow.

  \end{abstract}

%%%%%%%%% BODY TEXT

\section{Introduction}
Classification in high dimension requires building a representation that reduces a lot of variability while being discriminative. For example, in the case of images, there are geometric variabilities such as affine roto-translation, scaling changes, color changes, lighting, or intra-class variabilities such as style. Deep networks have been shown to be covariant to such actions \cite{aubry2015understanding,lenc2015understanding}, to linearize them \cite{radford2015unsupervised,nagamine2015exploring,zeiler2014visualizing} and combing those strategies permit building invariants to them \cite{bruna2013invariant,mallat2016understanding,fawzi2015manitest}. The creation of those invariants corresponds to a contraction of feature space, while separating the different classes. Convolutional Neural Networks (CNNs) consist of cascades of convolutional operators and non-linearities, and lead to state-of-the-art results on benchmarks where enough data are available \cite{krizhevsky2012imagenet,he2015deep}.

Geometric variabilities can be successfully handled by prior representations such as scattering transforms \cite{oyallon2015deep,bruna2013learning,mallat2016understanding,mallat2012group,sifre2013rotation}. They use cascades of wavelet transform and complex modulus to build a representation which is covariant to the action of several  groups of variabilities. Invariants to those variabilities are built by linear averaging that does not require to be learned, which corresponds to a contraction of the space along the orbits of those groups. Such priors can be derived from the physical laws that permit generating natural signals, such as the euclidean group. They lead to state-of-the-art results when all the groups of variabilities of the problem are understood, since in this case one only has a problem of variance. However, in the case of complex images datasets such as CIFAR10, CIFAR100 or ImageNet, understanding the nature of the non-geometrical invariants that are built remains an open question.

Several works address the problem to understand theoretically CNNs, but most of them assume the existence of a low-dimensional structure \cite{sokolic2016robust}. As in \cite{nagamine2015exploring}, we chose instead to have an empirical analysis. Our objective is to determine general properties which were not predictable by the theory and to characterize them rigorously. Are there restrictions on the property of a non-linearity? How can one  estimate a decision boundary? Can we find layerwise properties built by the deep cascade? Is there a layerwise dimensionality reduction?

In a classification task, one builds an estimator of the class for a test set using the samples of a training dataset. The intrinsic dimension of the classes is a quite low-dimensional structure in comparison with the original dimension of the signals, thus a classification task requires estimating a (often non-linear) projection onto a low-dimensional space. Observe that building this estimator is equal to estimating the boundary of classification of this task. It can be done in several ways that are related to each others. First, one can estimate the group of symmetry of the classification task as suggested in \cite{mallat2016understanding,bruna2013learning}. Those papers suggest that a deep network could potentially build a linear approximation of the group of symmetry of a classification task. For example, in the case of a CNN, all the layers are covariant with the action of translation which is a linear sub-group of symmetry of a classification problem, and this variability should be reduced. However, the class should not collapse when an averaging in translation is performed since they would become indistiguishable and lead to classification errors, thus \cite{mallat2016understanding} proposes to introduce  a notion of support vectors.  They are vectors of different classes,  which prevent the collapsing of different classes by indicating to the algorithm that it should carefully separate the different classes at those points. It means the deep network should not contract this part of the space, and they should build a classification boundary. Secondly, one could use the smoothness of the data. It means for example building an intermediate representation of the data that projects them  into a lower dimensional structure, for which building estimators of the class is simpler. For instance, it can be potentially done by linearizing high dimensional symmetries and then applying a projection. 

Understanding the nature of the classification boundary in the deep learning framework is especially hard, because the non-linear modules used are increasingly more complex, while improving progressively the benchmarks. For example, max pooling \cite{jarrett2009best}, spatial transformers \cite{jaderberg2015spatial}, local contrast normalization \cite{krizhevsky2012imagenet}, attention networks \cite{denil2012learning}, resnet \cite{he2015deep}, make the mathematical analysis even more difficult, since the way they are combined is mainly defined by a complex engineering process of trial and error based on the final accuracy of the network. However, \cite{springenberg2014striving} showed that a simple cascade of convolutions and ReLUs is enough to achieve good performances on standard datasets. The question of simplicity of a deep network is raised: what does simple mean? How simple can a deep network be, while leading to state-of-the-arts results?

Section \ref{simple} describes our architecture that depends upon few hyper parameters but leads to excellent numerical performances. Secondly, we discuss the variation of the hyper parameters of our architectures. Then,  Section \ref{cont} shows that the representation built by a deep network is progressively more regular. Finally,  we introduce a notion of \textit{local support vectors} which avoid the collapsing of classes, in Section \ref{sup}.  All experiments are reproducible using TensorFlow, via a software that is available online at: \url{https://github.com/edouardoyallon/deep\_separation\_contraction/}.

\section{A sandbox to understand deep networks} \label{simple}

We build a class of CNNs that depends on two hyper-parameters: its width and a non-linearity. We demonstrate that this framework is flexible and simple. First, we describe the setting that permits our network to reach the state of the art. Then, we vary those two hyper-parameters and observe counter-intuitive properties: a non-linearity does not need to be contractive, nor point wise, and a wide deep network generalizes better than a tight one.

\subsection{A spare pipeline for evaluation}
\begin{figure}[t]
\begin{center}
  
\begin{tikzpicture}[draw=black!50,y=-1cm]
\def \t{4}
\node at (0,0) [minimum width=2cm,line width=0] (x) {Input};
\node at (0,0.7) [minimum width=1.6cm,draw,align=center] (W1) {\small $\mathcal{B}_3$};
\node at (0,1.4) [minimum width=1.6cm,draw,align=center] (W2) {\small$\mathcal{B}_K$};
\node at (0,2) [] (W2n) {...};

\node at (-1.25,1.6) [,draw=none,line width=0] (p) {$5 \times$};
\draw (-0.75,1.1) -- (-1,1.1);
\draw (-1,1.1) -- (-1,2.1);
\draw (-1,2.1) -- (-0.75,2.1);

\node at (0,2.6) [minimum width=0.5cm,draw,circle,align=center,inner sep=1.5pt] (D1) {\small $\downarrow2$};

\def \u{1.9}
\node at (0,\u+1.4) [minimum width=1.6cm,draw,align=center] (W3) {\small$\mathcal{B}_K$};
\node at (0,\u+2) [] (W3n) {...};

\node at (-1.25,\u+1.6) [,draw=none,line width=0] (p) {$4 \times$};
\draw (-0.75,\u+1.1) -- (-1,\u+1.1);
\draw (-1,\u+1.1) -- (-1,\u+2.1);
\draw (-1,\u+2.1) -- (-0.75,\u+2.1);

\node at (0,\u+2.6) [minimum width=0.5cm,draw,circle,align=center,inner sep=1.5pt] (D2) {\small $\downarrow2$};

\def \v{3.9}
\node at (0,\v+1.4) [minimum width=1.6cm,draw,align=center] (W4) {\small$\mathcal{B}_K$};
\node at (0,\v+2) [] (W4n) {...};

\node at (-1.25,\v+1.6) [,draw=none,line width=0] (p) {$3 \times$};
\draw (-0.75,\v+1.1) -- (-1,\v+1.1);
\draw (-1,\v+1.1) -- (-1,\v+2.1);
\draw (-1,\v+2.1) -- (-0.75,\v+2.1);

\node at (0,6.4) [minimum width=1.6cm,draw,align=center] (A) {\small$A$};
\node at (0,7.1) [minimum width=1.6cm,draw,align=center] (L) {\small$L$};

\node at (0,7.8) [minimum width=2cm,line width=0] (fx) {Output};

\def \z{1}
\node at (0+\t,1.1+\z) [minimum width=2cm,line width=0] (xn) {};

\node at (-1.2+\t,1.4+\z) [minimum width=2cm,line width=0] (xnd) {$x_n$};
\node at (-1.2+\t,4.1+\z) [minimum width=2cm,line width=0] (xnd1) {$x_{n+1}$};
\draw[dashed,gray] (-0.95+\t,1.4+\z) -- (\t,1.4+\z);
\draw[dashed,gray] (-0.85+\t,4.1+\z) -- (\t,4.1+\z);

\draw [color=gray,thick](-2+\t,1+\z) rectangle (2+\t,4.5+\z);
\node at (\t,5+\z) [minimum width=2cm,line width=0] (xnd) {A block $\mathcal{B}_l$};

\node at (-0.7+\t,2.6+\z) [minimum width=2cm,line width=0] (xnd) {\tiny $3\times 3$};
\node at (-0.7+\t,2.8+\z) [minimum width=2cm,line width=0] (xnd) {\tiny $K\times l$};

\node at (0+\t,2+\z) [draw,circle,inner sep=0] (m) {$-$};

\node at (0.3+\t,1.9+\z) [] (m1) {\huge -};
\node at (-0.25+\t,1.75+\z) [] (m1) {\large +};

\node at (1.2+\t,2+\z) [draw] (Exn) {$Ex_n$};

\node at (0+\t,2.7+\z) [draw] (Wn) {$W_n$};

\node at (0+\t,3.5+\z) [draw] (rho) {$\rho$};
\node at (0+\t,4.4+\z) [] (o) {};

\draw[->,thick,black] (xn) -- (m);
\draw[->,thick,black] (Exn) -- (m);
\draw[->,thick,black] (m) -- (Wn);
\draw[->,thick,black] (Wn) -- (rho);
\draw[->,thick,black] (rho) -- (o);

\draw[->,thick,black] (x) -- (W1);
\draw[->,thick,black] (W1) -- (W2);
\draw[->,thick,black] (W2) -- (W2n);
\draw[->,thick,black] (W2n) -- (D1);
\draw[->,thick,black] (D1) -- (W3);
\draw[->,thick,black] (W3) -- (W3n);
\draw[->,thick,black] (W3n) -- (D2);
\draw[->,thick,black] (D2) -- (W4);
\draw[->,thick,black] (W4) -- (W4n);
\draw[->,thick,black] (W4n) -- (A);
\draw[->,thick,black] (A) -- (L);
\draw[->,thick,black] (L) -- (fx);
%\draw[->,thick,black] (P) -- (O);
%

\end{tikzpicture}
\end{center}
   \caption{Schematic representation of our architecture. Our network is a cascade of block $\mathcal{B}_l$, $l$ being the input size of the convolutional operator, followed by an averaging $A$ and a projection $L$.}
\label{fig:archi}
\end{figure}
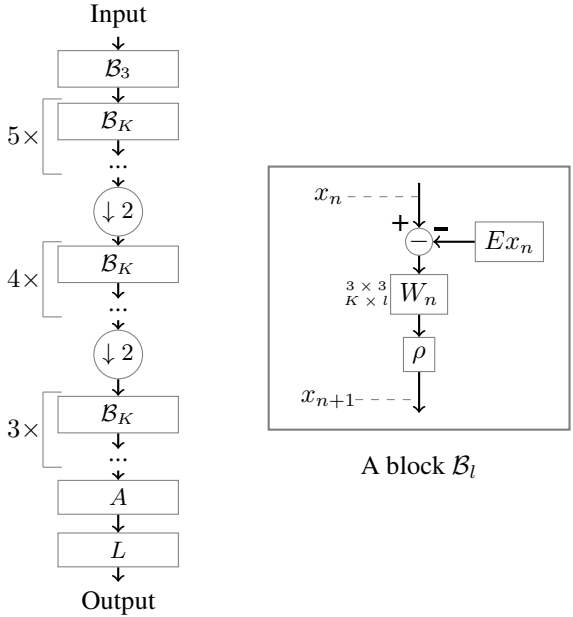

We describe the architecture that we used during all our experiment, with the datasets CIFAR10 and CIFAR100. It will depend only on $K\in \mathbb{N}$, which is the width of our network, and $\rho$ a non-linear function. Our deep network consists of the cascade of $13$ convolutional layers $W_n$ with non-linearity $\rho$. The spatial support of the kernel is $3\times 3$, and except for the first layer, the number of input and output layers is fixed equal to $K$. The output of the final convolutional layer is linearly and globally spatially averaged by $A$, and then reduced to the number of classes of the problem by a projection $L$. We did not learn any biases in the convolutional layers, however we subtract the mean $Ex_n$ from our feature maps, which is estimated on all the dataset via the standard batch normalization technique \cite{ioffe2015batch}. In this case we are in a similar setting as  \cite{mallat2013deep}, which proves that if $W_n$ is unitary then for any depths the network preserves the energy of the input signal and is non-expansive. For computational speed-up, we apply a spatial stride of $2$ at the output of the layers 6 and 10. Figure \ref{fig:archi} describes our network, which can be formally summarized for an input $x$, via $x_0=x$, and:
\[x_{n+1}=\rho W_n( x_n-Ex_n)\]
We trained our network via a SGD with momentum 0.9 to minimize the standard negative cross-entropy. We used a batch size of 128, and the training lasts 120 000 iterations. We used an initial learning rate of 0.25, that we divided by two every 10 000 iterations. To avoid overfitting, we apply 4 regularizations. First, a weight decay of 0.0002 that corresponds to a $l^2$ regularization. Then, we used dropout every two layers, starting at the second layer, that randomly sets  $40\%$ of the coefficients to 0: this is our main trick to achieve good performances. Thirdly, we used spatial batch normalization regularization that is supposed to remove instabilities during the training, as developed in  \cite{ioffe2015batch}. Finally, we applied standard random flipping and cropping techniques as data augmentation. Observe that we did not use any bias, simply removing the mean and did not use any non-linear pooling. Our architecture is thus kept as simple as possible, as in \cite{springenberg2014striving} but it only depends only on a few hyper parameters: its width and the non-linearity. Without any contrary mentions, we used $\rho=\text{ReLU}$ since it has heuristically been shown to achieve better performances. The first layer will always have a $\text{ReLU}$ non-linearity.

CIFAR10 and CIFAR100 are two  datasets of colored images of size $32\times 32$. The training set consists of 50 000 images, that are separated into 10 and 100 balanced classes respectively for CIFAR10 and CIFAR100. The testing set consists in 10 000 images. Those datasets are preprocessed using a standard procedure of whitening.

The number of parameters used by our network with CIFAR10 is $9\times (3K+12K^2)+10K$. To get our best accuracy, we used $K=512$ which corresponds roughly to 28M parameters, that lead to $95.4\%$ and $79.6\%$ accuracies on CIFAR10 and CIFAR100 respectively, which is an excellent performance according to Table \ref{tab:perf}. Thus, we are in a state-of-the-art setting to perform an analysis of the features learned.

\begin{table}

\begin{center}
\begin{tabular}{|l|c|c|c|c|}
\hline
Methods&Depth&\#params&\small CIFAR10&\small CIFAR100\\
\hline\hline
\textbf{Ours} & 13&28M &95.4 &79.6\\
SGDR \cite{loshchilov2016sgdr}  &28 &150M &96.2&82.3\\
RoR \cite{zhang2016residual} &58 &13M&96.2 &80.3\\
WResNet \cite{zagoruyko2016wide} & 28 &37M &95.8 &80.0\\
All-CNN \cite{springenberg2014striving} &9 &1.3M&92.8 &66.3\\

\hline

\end{tabular}

\end{center}

\caption{Accuracy on CIFAR10 and CIFAR100 for state-of-the-arts supervised deep networks. Depth and number of parameters are reported to perform a fair comparison.}

\label{tab:perf}
\end{table}

\subsection{Weakening the non-linearity}
Contraction phenomenon is a necessary step to explain the tremendous dimensionality reduction of the space that occurs. A network cannot be purely linear, since some classification problems are not linearly separated: indeed a linear operator can only contract along straight lines. Should $\rho$ also be a contracting operator?  We study specifically the pointwise non linearity $\rho$ in a CNN and its necessary conditions to reach good classification accuracy. 

\subsubsection{Unneccesity to contract via $\rho$}
Since the AlexNet \cite{krizhevsky2012imagenet}, non-linearity is often chosen to be a $\text{ReLU}(x)=\max(0,x)$. This is a non-expansive function, e.g. $|\text{ReLU}(x)-\text{ReLU}(y)|\leq |x-y|$, and also continuous. Consequently, the cascade of linear operators of norm less than 1 and this non-linearity is non-expansive which is a convenient property to the reduce or maintain the volume of the data.

Modulus non-linearity in complex network have been also suggested to remove the phase of a signal, which corresponds in several frameworks to a translation variability \cite{mallat2012group,bruna2013invariant} . For instance, if the linear operator consists in a wavelet transform with appropriate mother wavelet \cite{mallat1999wavelet}, then the spectrum of each convolution with a wavelet is localized in Fourier. This implies that a  small enough translation in the spatial domain will also result in a phase multiplication in the spatial domain. Applying a modulus removes this variability. As a classical result of signal theory, observe also that an averaged rectified signal is approximately equal to the average of its complex envelope \cite{mallat1999wavelet}. Consequently, cascaded with an average pooling, a ReLU and a modulus might have the same use.

Experimentally, it is possible to build a deep network that leads to $89.0\%$ accuracy on CIFAR10, with $K=256$, with the non-linearity chosen as:

\[\rho(x)=\text{sign}(x)(\sqrt{|x|}+0.1)\]
In the neighborhood of 0, this non-linearity is not continuous in 0, has an arbitrary large derivative, and preserves the sign of the signal. It shows that continuity property, lipschitz property or removing the phase of the signals are not necessary conditions to obtain a good accuracy. It suggests that more refinement in the mathematical analysis of $\rho$ is required.

\subsubsection{Degree of non-linearity}
In this subsection, we try to weaken the traditional property of pointwise non-linearity. Indeed, being non-linear is essential to ensure that the different classes can be separated, however the recent work on ResNet \cite{he2015deep} suggests that it is not necessary to apply a pointwise non-linearity, thanks to identity mapping that can be interpreted as the the concatenation of a linear block (the identity) and a non-linear block. In this case, a non-linearity is applied only on a half of the feature maps. We investigate the question to understand if this property generalizes to our architecture by introducing a ReLU with a degree $\frac k K$ of non-linearity that we apply to a feature map $x(u,l)$, where $u$ is the spatial variable and $l$ the index of the feature map, defined by:
 \[\text{ReLU}^K_k(x)(u,l)\triangleq\begin{cases}\text{ReLU}(x(u,l)), & \text{if }l \leq k   \\ x(u,l), & \text{otherwise}\end{cases}\]

\begin{figure}[t]
\begin{center}
   \includegraphics[width=0.95\linewidth]{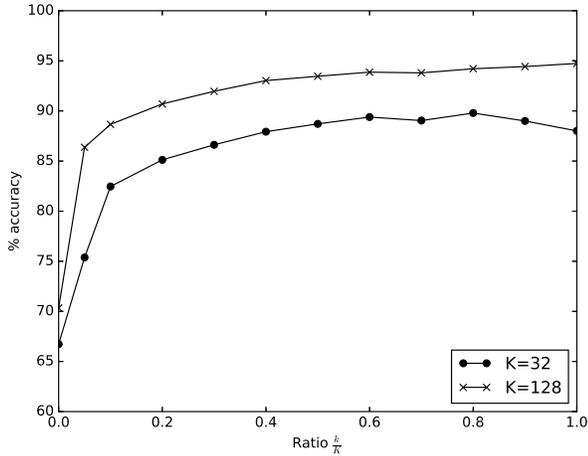}
\end{center}
   \caption{Accuracy  when varying the degree of non-linearity $\frac{k}{K}$, reported with  $K=32$ and $K=128$. When $k=K$, one obtains $88.0\%$ and $94.7\%$ respectively for $K=32$ and $K=128$. The maximum accuracies are then respectively 89.8\% and 94.7\%, which indicates that a point-wise non-linearity is not necessary the optimal configuration.}
\label{fig:deg}
\end{figure}

In the case $k=0$, we have an almost  linear network (there is the ReLU non-linearity at the first layer), and when $k=K$, it is a standard deepnetwork with point-wise non-linearity. Figure \ref{fig:deg} reports the numerical accuracy when we vary $k$, fixing $K$ equal to $32$ or $128$. A linear deep network performs poorly, leading to an accuracy of roughly $70\%$ on CIFAR10. We see that there is a plateau when $\frac{k}{K}\geq 0.6=\frac{k_0}{K}$, and that the maximum accuracy is not necessary obtained for $k=K$. Our networks could reach $89.8\%$ and $94.4\%$ classification accuracy respectively for $K=32$ and $K=128$.

This is an opportunity to reinterpret the non-linearity. Let $\tau$ be a cyclic translation of $\{1,...,K\}$, e.g. $\tau([1,...,K])=[K,1,...,K-1]$, such that we define: $\tau(x)(u,l)\triangleq x(u,\tau(l))$. In this case, $\tau$ is a linear operator that translates cyclically the channels of a feature map. Observe that:
 \[\RELU{K}{k}x=\underbrace{\tau \circ \RELU{K}{1}\circ... \tau \circ \RELU{K}{1}}_{k\text{ times}}x\]
In this setting, one might interpret a CNN with depth $N$ and width $K$  as a CNN of depth $NK$ and width $K$, since it is also a cascade of $NK$ $\RELU{K}{1}$ non-linearities and $\{\tau,W_n\}_n$ linear operators. In this work, $\tau$ is fixed, yet it might be learned as well. It means also that if $k<K$, by increasing the number of layers, a CNN using a ReLU non-linearity can be rewritten with a $\RELU{k}{K}$ non-linearity. For $k<k_0$, we tried to increase the depth of the deep network to recover its best accuracy since there will be as much non-linearity as in the case $k=K$, and we know there exists an optimal solution. However, our network was not able to perform as well, which implies that there is an issue with the optimization. Restricting the non-linearity application to only one feature map could help future analysis, since it gives explicitly the coefficients that exhibits non-linear effects. Finally, the only hyper parameter that remains is the number of feature maps $K$ of the layers, that we study in the next final subsection.

\subsection{Increasing the width}\label{params}
\begin{figure}[t]
\begin{center}
   \includegraphics[width=0.95\linewidth]{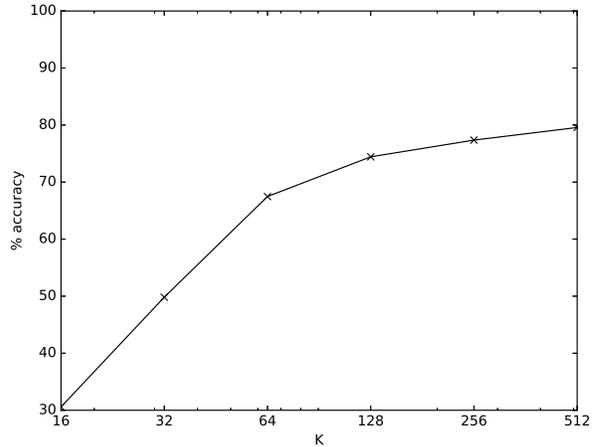}
\end{center}
   \caption{Accuracy when varying $K$ on CIFAR100 dataset, the axis of $K$ is in log scale.}
\label{fig:c10}
\end{figure}

\begin{figure}[t]
\begin{center}
   \includegraphics[width=0.95\linewidth]{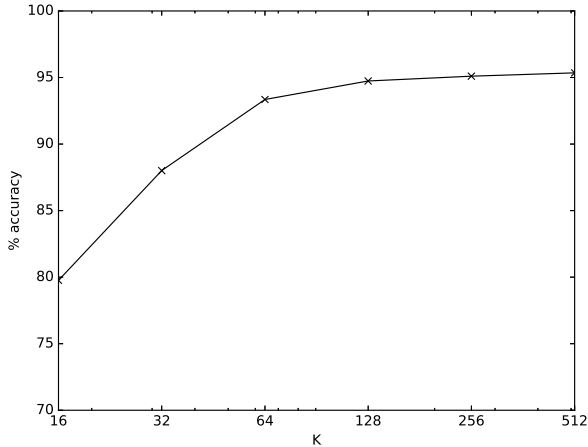}
\end{center}
   \caption{Accuracy when varying $K$ on CIFAR10 dataset, the axis of $K$ is in log scale.}
\label{fig:c100}
\end{figure}

In this section, we show that increasing $K$ increases the classification accuracy of the network. \cite{zagoruyko2016wide} reports also this observation, which is not obvious since increasing $K$ increases by $K^2$ the number of parameters and could lead to a severe overfitting. Besides, since a final dimensionality reduction must occur at the last layer, one could expect that the intermediate layers might have a small number of feature maps. Figure \ref{fig:c10} and  \ref{fig:c100} reports the numerical accuracy respectively on CIFAR10 and CIFAR100, with respect to $K$. Setting $K=512$ leads to $95.4\%$ and $79.6\%$ accuracy respectively on CIFAR10 and CIFAR100, while $K=16$ leads to $79.8\%$ and $30.6\%$ accuracy on CIFAR10 and CIFAR100 respectively. It is not clear if the reason of this improvement is the optimization or if it is a structural reason. Nevertheless, it indicates that increasing the number of feature maps is a simple way to improve the network accuracy.


\section{Contracting the space} \label{cont}

Regularity of a representation with respect to the classes is necessary to classify high-dimensional samples. Regularity means here that a supervised classifier builds a covering of the full data space with training samples via $\varepsilon$-balls that is small in term of volume or number of balls, yet that it still generalizes well \cite{vapnik2000methods}. The issue is that it is hard to track this measure. For example, assume the data lay on a 2D or 3D manifold and are smooth with respect to the classes, then it is possible to build a classifier which is locally affine using manifold learning techniques. In particular, this implies that the euclidean metric is locally meaningful. We take a second example: when a nearest neighbor obtains good generalization properties on a test set. In this case, our problem is regular since it means again that locally the euclidean metric is meaningful and that the representation of the training set is mostly covering the data space. We show that supervised deep networks progressively build a representation where euclidean distance becomes more meaningful. Indeed, we numerically demonstrate that the performance of local classifiers, e.g. which assign a class by giving more important weights to points of the training set that are in a neighborhood of the testing sample, progressively improves with depth.

In this section and the following, in order to save computation time, we used the network previously introduced with $K=32$ and $\rho = \text{ReLU}$. Numerically, we however tried several parameters as a sanity check that our conclusion generalizes to any values of $K$, to avoid a loss in generality. With $K=32$, the accuracy of the network is $88.0\%$. Increasing $K$ to 512 should approximatively increase all the reported results by $7$ absolute percents of accuracy.

Translation is one of the symmetries of the image classification problem, thus it is necessary to remove this variability, even if the features are not extracted at the final layer. In the following, we perform our experiments using $\bar x\triangleq Ax \in \mathbb{R}^{32}$. We have the following $\epsilon$ separation property, thanks to non-expansivity of averaging operators: 
\begin{equation}
\Vert x-y\Vert \geq \Vert \bar x-\bar y \Vert \geq \epsilon
\label{eq:1}
\end{equation}
 It means that a separation by $\epsilon$ of the averaged signals implies a separation by at least $\epsilon$ of any translated versions of the original signals. We  denote the features at depth $n$ of the training set by $\Xa{n}=\{\bar x_n, x\in \text{training}\}$ and the features of depth $n$ of the testing set by $\Xe{n}=\{\bar x_n, x\in \text{testing}\}$. For a given $x$, $x_n$ or $\bar x_n$, we write its class $y(x)$, $y(x_n)$ or $y(\bar x_n)$ since there is no confusion possible.

\subsection{A progressively more regular representation}\label{prog}
\begin{figure}[t]
\begin{center}
   \includegraphics[width=0.95\linewidth]{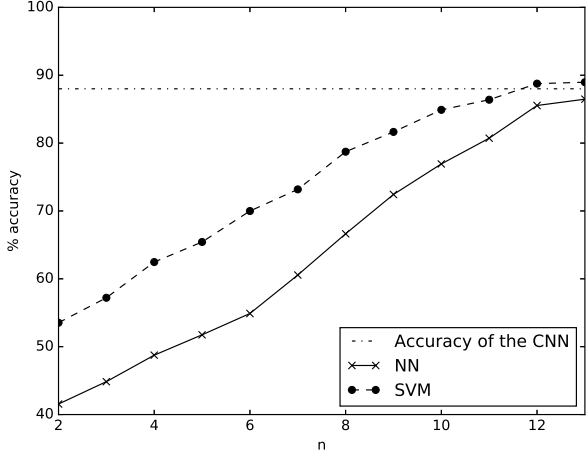}
   \caption{Accuracy on CIFAR10 at depth $n$ via a Gaussian SVM and 1-NN. The size of the network is $K=32$ and its accuracy on the testing set is $88.0\%$.}
\label{fig:acc}
\end{center}

\end{figure}

Brutal contraction via a linear projection of the space would not preserve the distances between different classes. Yet, with a cascade of linear and non linear operators, it would be possible to progressively contract the space without losing in discriminability \cite{mallat2013deep}. Several works \cite{zeiler2014visualizing,nagamine2015exploring} reported that linear separability of deep features increases with depth. It might (but this is not the only solution) indicate that intra-class variabilities are progressively linearized \cite{mallat2016understanding}, until the last layer, such that a final linear projector can build invariants. Following this approach, we start by applying a Gaussian SVM at each depth $n$, which is a discriminative locally linear classifier, with a fixed  bandwith. Indeed, observe here that the case of an infinite bandwidth corresponds to a linear SVM. We  train it on  the standardized features $\Xa{n}$ corresponding to the training set at depth $n$ and test it on $\Xe{n}$, via a Gaussian SVM with bandwith equal to the average $l^2$ norm of the points of the training set. We only  cross-validate once the  regularization parameter at one layer and then kept the parameters of the SVM constant. Figure \ref{fig:acc} reports that the accuracy of this classifier increases at regular step with depth, which confirms the features become more  separable.

In fact, we prove this Gaussian SVM acts as a local classifier. A 1 nearest neighbor (1-NN) classifier  is a naive and simple non-parametric classifier for high-dimensional signals, that simply assigns to a point the class of  its closest neighbor. It can be interpreted as a local classifier with adaptive bandwith \cite{burman1992location}. It is unbiased, yet it bears a lot of  variance. Besides, resampling the data will affect a lot the classification results. We train a 1-NN on $\Xa{n}$ and test it on $\Xe{n}$. We denote by a $\NN{k}{x}$ the result of the $k$-th closest neighbor of a point, that is distinct of itself. We  observe in Figure \ref{fig:acc} that a 1-NN trained on $\Xa{n}$ and tested on $\Xe{n}$ performs nearly as well as a Gaussian SVM. Besides, the progression of this classifier is almost linear with respect to the depth.  It means that the representation built by a deep network is in fact progressively more regular, which explains why the Gaussian SVM accuracy progressively improves.

\subsection{A progressive reduction of the space}
\begin{figure}[t]
\begin{center}
   \includegraphics[width=0.95\linewidth]{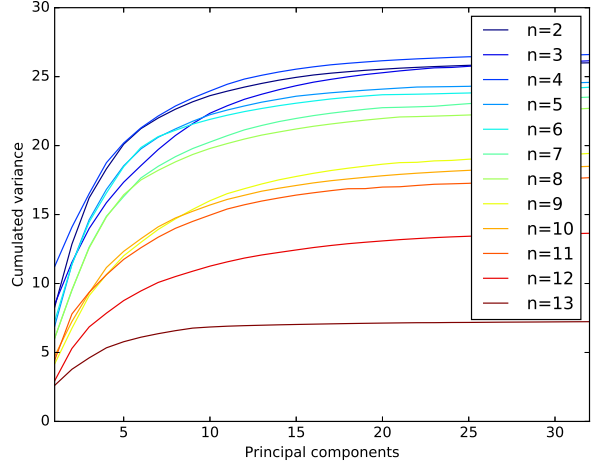}
\end{center}
   \caption{Cumulated variances of the principal component of a given class at different depths $n$, for a network trained on CIFAR10 with $K=32$. In general, one observes a reduction of variance with depth. Best viewed in color.}
\label{fig:PCA}
\end{figure}

We want to quantify   the contraction of the space performed by our CNN. In this subsection, we show that the samples of a same class define a structure that progressively becomes low-dimensional. We investigate the question to understand if the progressive improvement of the 1-NN is due to a dimensionality reduction. First, we check wether a linear dimensionality reduction is implemented by our CNN.  To this end, we apply a PCA on the  features $\Xa{n}$ belonging to the same class at each depth $n$. As a normalization, the features at each depth $n$ were globally standardized.  In other words, the data at each depth are in the $l^2$ balls of radius 32.  Remember that in our case, $\bar{x}_n\in \mathbb{R}^{32}$. Figure \ref{fig:PCA} represents the cumulated variances of the $K=32$ principal component axis of a given class at different depth $n$. The obtained diagram and conclusions are not specific to this class. The accumulated variance indicates the proportion of energy that is explained by the first axis. The slope of a curve is an indicator of the dimension of the classes: as a plateau is reached, the last components are not useful to represent the class for classifiers based on $l^2$ distances. The first observation is that the variance seems to be uniformly reduced with depth. However, certain plots of successive depths are almost indisguishable: it indicates that almost no variance reduction has been performed. Except for the last layer, the decay of the 20 last eigenvalues is slow: this is not surprising since nothing indicates that the dimension should be reduced and small variance coefficients might be  important for the classification task. The  very last layer exhibits a large variance reduction and is low-dimensional, yet this is logical since by construction, the final features should be linearly separable and be in space of dimension 10.

\begin{figure}[t]
\begin{center}
   \includegraphics[width=0.95\linewidth]{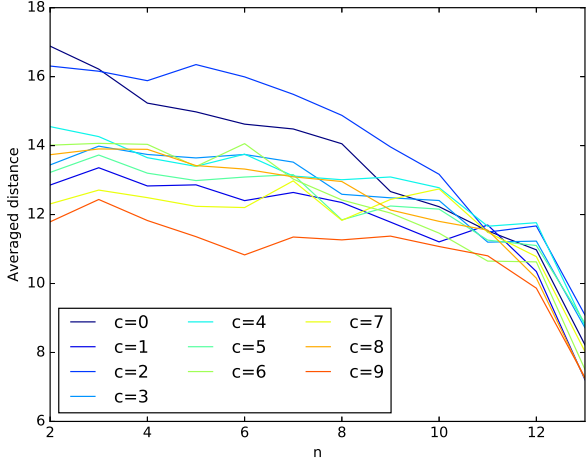}
\end{center}
   \caption{Averaged intra-class distances on CIFAR10, $K=32$, at different depths $n$. Different colors correspond to different classes. The intra-class distances are globally decreasing with depth.  Best viewed in color.}
\label{fig:intra}
\end{figure}

We then focus on the contraction of the intra-class distances. As a cascade of non-expansive operators, a deep network is also non-expansive up to a multiplicative constant. In particular, the intra-class distances should be smaller. In Subsection \ref{prog}, we observed that the euclidean metric was in fact meaningful: this could indicate low-dimensional manifold structure. Under this hypothesis, in a similar fashion as \cite{giryes2015deep}, we study the average intra-class distances. As a normalization, the features $\Xa{n}$ at depth $n$ are standardized over the dataset. On CIFAR10 (where each of the 10 classes has 5000 samples) we compute an estimation of the average distances of the intra-class samples of the  features $\Xa{n}$ at depth $n$ for the class $c$:
\[\frac{1}{5000^2}\sum_{\substack{\bar{x}_n \in \Xa{n}\\ y(x_n)=c}}\sum_{\substack{\bar{x'}_n \in \Xa{n}\\ y( x'_n)=c}}\Vert \bar{x}_n-\bar{x'}_n\Vert  \]
Figure \ref{fig:intra} reports this value for different classes $c$ and different depths $n$. One sees that the intra-class distances do not strictly decrease with depth, except on the last layer, which must be low-dimensional since the features are, up to projection, in a space of size 10. This is due to two phenomena: the normalization procedure whose choice can drastically change the final results and the averaging. Indeed, let us assume here that $\Vert W_n \Vert \leq 1$, then if $x,\tilde x$ are in the same class, $\Vert x_{n+1} - \tilde x_{n+1}\Vert \leq  \Vert x_{n} - \tilde x_{n}\Vert$, but this does not imply that $\Vert \bar x_{n+1} - \bar{x'}_{n+1}\Vert \leq  \Vert \bar x_{n} - \bar{x'}_{n}\Vert$, since the averaging is a projection and could break the distance inequality.

Those two experiments indicate we need to refine our measurement of contraction to explain the progressive and constant improvement of a 1-NN. Specifically, one should estimate the local intrinsic dimension:  this is not possible since we do not have enough available samples in each neighborhood \cite{costa2005estimating}.

\section{Local support vectors for  building  a variable bandwidth classification boundary}\label{sup}

Our objective is to quantify at each depth $n$, the regularity of the representation constructed by a deep net in order to understand the progressive contraction of the space. In other words, we need to build a measure of this regularity.   The contraction of the space is global, but we know from below that neighbors are meaningful: we woud like to explain how they separate the different classes.  We thus introduce a notion of \textit{local support vectors}. In the case of a SVM, a support vector corresponds to samples of the training set that delimit different classes, by interpolating a hyperplane between them \cite{cortes1995support}. It means that a support vectors permits to avoid the collapsing of the boundary classification \cite{mallat2016understanding}. But in our case, we do not have enough samples to estimate the exact boundaries.  \textit{Local support vectors} corresponds to support vectors defined by a local neighborhood. In other words, at depth $n$, the set of support vectors is defined as $\Gamma_n=\{\bar{x}_n,y(\NN{1}{\bar{x}_n})\neq y(\bar{x}_n)\}\subset \Xa{n}$ which is the set of nearest neighbors that have a different class. In this section for a finite set $X$, we denote its cardinality by $|X|$.

\subsection{Margin}
In this subsection, we numerically observe a margin between support vectors. In \cite{sokolic2016robust}, bounds on the margin are obtained with hypothesis of low dimensional structures, and this might be restrictive according to the analysis above. A margin at depth $n$ is defined as:
\[\gamma^n=\inf_{y(\NN{1}{\bar x_n})\neq y(\bar x_n)}\Vert\NN{1}{\bar x_n}-\bar x_n\Vert\geq 0\]

\begin{figure}[t]
\begin{center}
   \includegraphics[width=0.95\linewidth]{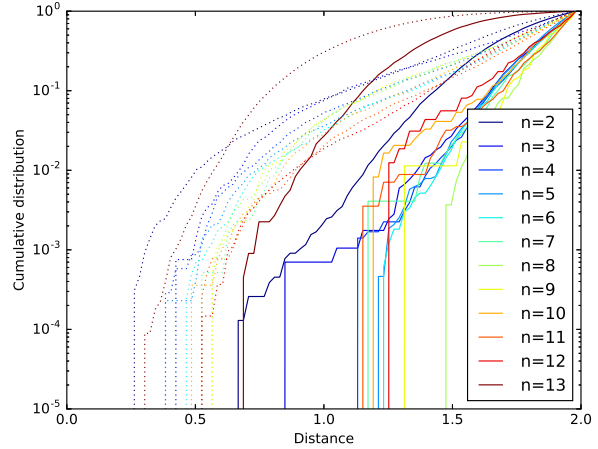}
\end{center}
   \caption{Cumulative distributions of distances: between a support vector and its nearest neighbors, e.g. $\mathcal{B}_n(t)$  (dashed line), and a point that is not a support vector and its nearest neighbor, e.g. $\mathcal{A}_n(t)$ (continuous line). Different colors correspond to different depths. The axis of the magnitude of the distance is in log scale. At a given depth, one sees there is a significantive difference between the cumulative distribution, which indicates the existence of a margin.  Best viewed in color.}
\label{fig:dist}
\end{figure}

Since our data are in finite number this quantity is always different from 0, but we need to measure if it is significant. We thus compare the distributions of distances of nearest neighbors belonging to the same class $A_n=\{\Vert\NN{1}{\bar{x}_n}-\bar{x}_n\Vert,\bar{x}_n\not \in \Gamma_n\}$ and the distributions of the distances between support vectors $B_n=\{\Vert\NN{1}{\bar{x}_n}-\bar{x}_n\Vert,\bar{x}_n \in \Gamma_n\}$. The features have been normalized by a standardization. Figure \ref{fig:dist} represents the cumulative distributions of $A_n$ and $B_n$ for different depths $n$. We recall that a cumulative distribution of a finite set $A\subset \mathbb{R}$ is defined as: $\mathcal{A}(t)=\frac{|\{x\leq t,x\in A\}|}{|A|}$.  One observes that in a neighborhood of 0,  $\mathcal{A}_n(t)$ is roughly the translation of   $\mathcal{B}_n(t)$ by 0.5. It indicates there is a significant difference, showing $\gamma_n$ is actually meaningful. Consequently there exists a margin between the spatially averaged samples of different classes, which means by Equation \eqref{eq:1} that  this margin exists between the samples themselves and their orbits by the action of translations.

\subsection{Complexity of the classification boundary}
\begin{figure}[t]
\begin{center}
   \includegraphics[width=0.95\linewidth]{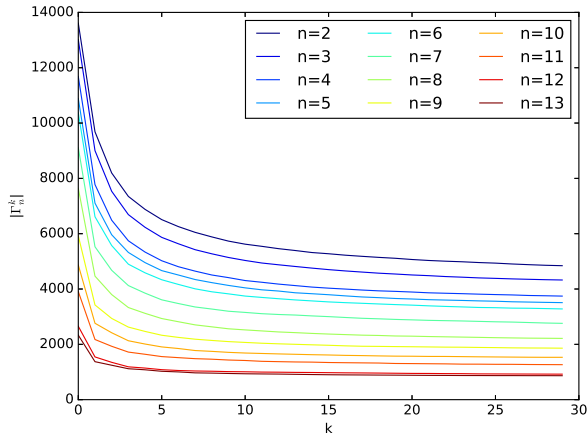}
\end{center}
   \caption{$|\Gamma^k_n|$ for different depth $n$. Different depths are represented by different colors. The limit value of $|\Gamma^k_n|$ is reached faster by deeper layers, and the value $|\Gamma^k_n|$ globally decrease with depth: the boundary classification is progressively more regular.  Best viewed in color.}
\label{fig:SV}
\end{figure}

Estimating a local intrisic dimension is difficult when few samples per neighborhood are available, but  the classes of the neighbors of the samples of $\Xa{}$ are known. In this subsection, we build a measure of the complexity of the classification boundary  based on neighbors. This permits evaluating both separation and contraction properties. It can be viewed as a weak estimation of the intrisic dimension \cite{brito2013intrinsic}, even if the manifold hypothesis might not hold. We compute an estimate of the efficiency of a k-NN to correctly find the label of a local support vector. To this end, we define by recurrence at depth $n$ and for a given $k\in \mathbb{N}$, which is a number of neighbors, $\Gamma^k_n$ via $\Gamma^1_n=\Gamma_n$ and:
\[\Gamma^{k+1}_n= \Bigl \{\bar{x}_n\in \Gamma^{k}_n, |\{y(\NN{l}{\bar{x}_n})\neq y(\bar{x}_n),l\leq k+1\}|>\frac{k}{2}\Bigr \}\]
In other words, $\Gamma_n^k$ is the set of points at depth $n$ that are not well-classified by  $l$-NNs using majority vote, for $l \leq k$.  By construction, $\Gamma^{k+1}_n\subset \Gamma^k_n$ which implies that $|\Gamma^{k+1}_n|\leq |\Gamma^k_n|$. Since the number of samples is finite, this sequence converges to  the number of samples of the training set that can not be identified by their nearest neighbors. The decay and the amplitude of $|\Gamma^k_n|$ is an indicator of the regularity of the classification boundary. Recall that for a deep network, the 1-NN classifier has better generalization properties with deeper features. A small value of $|\Gamma^k_n|$ indicates that a few samples are necessary to build the classification boundary (contraction), and at a given depth $n$, if $|\Gamma^k_n|$ decreases quickly to its constant value, it means a few neighbors are required to build the decision boundary (separation). Figure \ref{fig:SV} indicates that the classification boundary is uniformly more regular with depth, in term of number of local support vectors and number of neighbors required to estimate the correct class. This measure has the advantage of being simple to compute, yet this analysis must be refined in a future work.

\section{Conclusion}
In this work, we simplified a standard deep network that still reaches good accuracies on CIFAR10 and CIFAR100. We studied the influence of different hyperparameters such as the non-linearity and the number of feature maps. We demonstrate that the performance of a nearest neighbors classifier applied at different depth increases and that this classifier is almost as discriminative as a Gaussian SVM. Finally, we defined local  support vectors that allow us to build a  measure of the contraction and separation properties of the built representation. They could permit to potentially improving the classification accuracy, by refining the boundary classification of a CNN in their neighborhood.

We have built a class of deep networks that uses only pointwise non-linearities and convolutions and that should help for future analysis.  In a context where the use of deep networks is increasing impressively, understanding the nature of the intrisic regularity they exploit is mandatory.  Solving this problem will help finding theoretical guarantees on deep networks for applications and must be the topic of future research.

\ifcvprfinal{\section*{Acknowledgement} I would like to thank  Mathieu Andreux, Tom\'as Angles, Bogdan Cirstea, Michael Eickenberg and St\'ephane Mallat for helpful discussions and comments.  This work is funded by the ERC grant InvariantClass 320959 and via a grant for PhD Students of the Conseil r\'egional d'Ile-de-France (RDM-IdF).}\fi

{\small
\bibliographystyle{ieee}
\bibliography{paper}
}

\end{document}